# MRI Brain Tumor Segmentation using Random Forests and Fully Convolutional Networks

Mohammadreza Soltaninejad[1], Lei Zhang[1], Tryphon Lambrou[1], Guang Yang[2], Nigel Allinson[1], and Xujiong Ye[1]

[1]Laboratory of Vision Engineering, School of Computer Science, University of Lincoln, UK
[2]National Heart & Lung Institute, Imperial College London, UK
{msoltaninejad,lzhang,tlambrou,nallinson,xye}@lincoln.ac.uk,
g.yang@imperial.ac.uk

**Abstract.** In this paper, we propose a novel learning based method for automated segmentation of brain tumor in multimodal MRI images, which incorporates two sets of machine-learned and hand crafted features. Fully convolutional networks (FCN) forms the machine learned features and texton based features are considered as hand-crafted features. Random forest (RF) is used to classify the MRI image voxels into normal brain tissues and different parts of tumors, i.e. edema, necrosis and enhancing tumor. The method was evaluated on BRATS 2017 challenge dataset. The results show that the proposed method provides promising segmentations. The mean Dice overlap measure for automatic brain tumor segmentation against ground truth is 0.86, 0.78 and 0.66 for whole tumor, core and enhancing tumor, respectively.

**Keywords:** Fully Convolutional Networks, Random Forest, Deep Learning, Texton, MRI, Brain Tumor Segmentation.

## 1 Introduction

Accurate segmentation of brain tumor may aid the fast and objective measurement of tumor volume and also find patient-specific features that aid diagnosis and treatment planning [1]. Due to the recent advances in deep neural networks (DNN) in recognition of the patterns in the images, most of the recent tumor segmentations have focused on deep learning methods. Recently, fully convolutional networks (FCN) have been suggested for dense (i.e. per-pixel) classification with the advantage of end-to-end learning [2], without requiring those additional blocks in convolutional neural networks (CNN) based approaches. Despite the advantage of dense pixel classification, FCN-based methods still have limitations of considering the local dependencies in higher resolution (pixel) level. The loss of spatial information, which occurs in the pooling layers, results in coarse segmentation. This limitation will be addressed in our work by incorporating high resolution hand-crafted textural features which consider local dependencies of the pixel. Texton feature maps [3] provides significant information on multi-resolution image patterns in both spatial and frequency domains.

In this paper, a novel fully automatic learning based segmentation method is proposed, by applying hand-crafted and machine-learned features to random forest (RF) classifier. The machine-learned FCN based features detect the coarse region of the tumor while the hand-crafted texton descriptors consider the spatial features and local dependencies to improve the segmentation accuracy.

## 2 Methods

Our method is comprised of four major steps (pre-processing, FCN, Texton map generation, and RF classification) that are depicted in Fig. 1. In the pre-processing stage, the intensities were normalized for each protocol by subtracting the average of intensities of the image and dividing by their standard deviation. Then, the histogram of each image was normalized and matched to the one of the patient images which is selected as the reference.

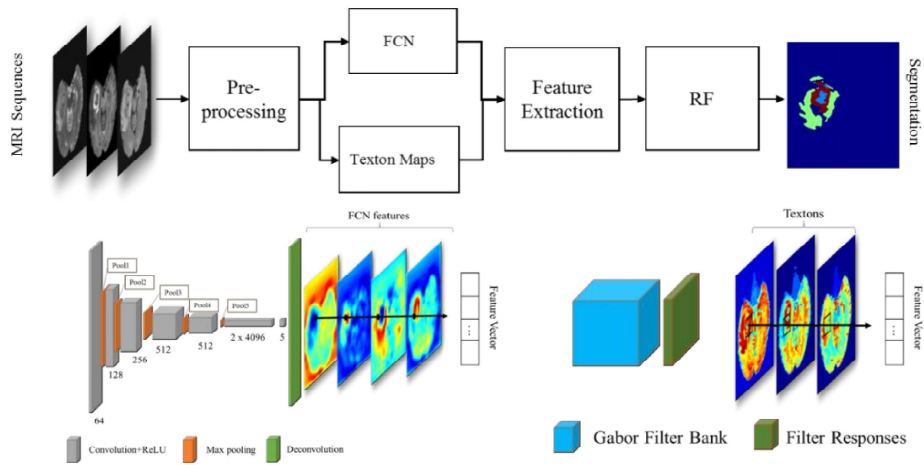

**Fig. 1.** Flowchart of the proposed method. The FCN architecture, machine learned feature extraction, and texton features.

### 2.1 FCN-based Features

A FCN-8s architecture in [2] was adopted for segmentation of brain tumor in multi-modal MRI images, where the VGG16 [4] was employed as CNN classification net. The FCN-8s was constructed from FCN-16s skip net and FCN-32s coarse net, which was implemented by fusing predictions of shallower layer (Pool3) with 2 × upsampling of the sum of two predictions derived from pool4 and last layer. Then the stride 8 predictions were upsampled back to the image.

The feature vector is generated for each voxel based on the score map from the FCN. For each class label, a score map is generated, 4 maps are generated using the standard BRATS17 labelling system. The values of each map layer corresponding to each voxel are considered as machine-learned features of that voxel.

### 2.2 Texton features

The texton based features were applied in the proposed method as hand-crafted features to support the machine-learned features and improve the segmentation results. Textons are obtained by convolving the image with a Gabor filter bank. To cover all orientations six different filter directions were used: [0º, 30º, 45º, 60º, 90º, 120º]. Filter sizes were [0.3, 0.6, 0.9, 1.2, 1.5] and the wavelength of sinusoid coefficients of the Gabor filters were [0.8, 1.0, 1.2, 1.5].

The MR images were convolved with the Gabor filters, then the filter responses are merged together and clustered using *k-means* clustering. The number of clusters 16 was selected as the optimum value for the number of clusters in texton map. The texton map is created by assigning the cluster number to each voxel of the image. The texton feature for each voxel is the histogram of textons in a neighborhood window of $5 \times 5$ around that voxel.

The normalized intensity value of the voxels in each modality which is obtained from the pre-processing stage is also included in the feature vector. Therefore, in total 55 features were collected (4 FCN score maps, 3 protocol intensities and 48 texton histograms) for usage in the next step.

### 2.3 RF Classification

The potential tumor area detected by the FCN output was considered as the initial region of interest (ROI). This ROI was selected as a confidence margin of 10 voxels in 3D space around the detected initial tumor area which was calculated by morphological dilation. The feature vectors for voxels in this ROI were fed to the random forests. The main parameters in designing RF are tree depth and the number of trees. RF parameters were tuned by examining different tree depths and number of trees on training datasets and evaluating the classification accuracy using 5-fold cross validation. The number of 50 trees with depth 15 provided an optimum generalization and accuracy. Based on the classes assigned for each voxel in the validation dataset, the final segmentation mask was created by mapping back the voxel estimated class to the segmentation mask volume. Finally, the bright regions in the healthy part of the brain near to the skull were eliminated using a connected component analysis.

## 3 Results

The proposed method was performed on MATLAB 2016b on a PC with CPU Intel Core i7 and RAM 16 GB with the operating system windows 8.1. The FCN was implemented using MatCovNet toolbox [5]. GPU GeForce gtx980i was used to run FCN. The RF was implemented using open source code provided in [6] which is a specialized toolbox for RF classification based on MATLAB.

Both FCN and RF are trained on BRATS 2017 [7–10] training dataset which include 220 high grade glioma (HGG) and 75 low grade glioma (LGG) patient cases. The method was evaluated on BRTAS 2017 validation dataset which include 46 patient cases.

The evaluation measure which are provided by the CBICA's Image Processing Portal, i.e. Dice score, sensitivity, specificity, Hausdorff distance, were used to compare the segmentation results with the gold standard (blind testing). Figure 2 shows segmentation results of the proposed method on some cases of BRATS 2017 validation dataset. Table 1 provides the evaluation results obtained by applying the proposed method on BRATS 2017 validation dataset.

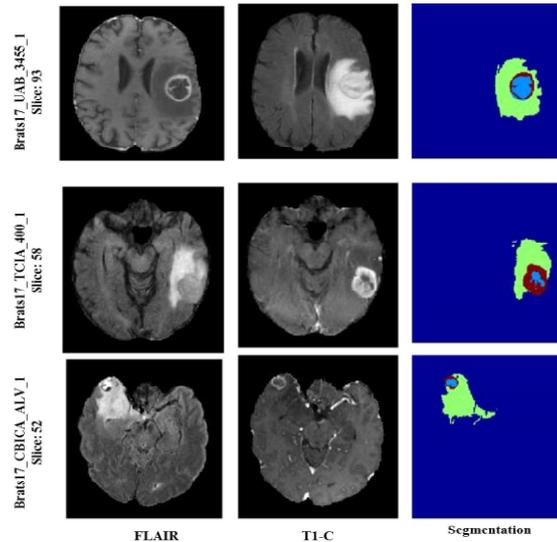

**Fig. 2.** segmentation masks for validation dates, using the proposed method. Light blue: necrosis and on-enhancing, green: edema, dark red: enhancing.

**Table 1.** Segmentation results for validation dataset which was provided by CBICA portal. ET: enhancing tumor, WT: whole tumor, TC: tumor core.

|      | Dice |      |      | Sensitivity |      |      | Specificity |      |      | Hausdorff (95%) |       |       |
|------|------|------|------|------|------|------|------|------|------|------|-------|-------|
|      | ET   | WT   | TC   | ET   | WT   | TC   | ET   | WT   | TC   | ET   | WT    | TC    |
| Mean | 0.66 | 0.86 | 0.78 | 0.57 | 0.83 | 0.72 | 1.00 | 1.00 | 1.00 | 3.76 | 7.61  | 8.70  |
| STD  | 0.28 | 0.09 | 0.19 | 0.28 | 0.13 | 0.21 | 0.00 | 0.01 | 0.00 | 4.38 | 12.99 | 13.52 |

## 4    Conclusion

In this paper, a novel method was proposed in which the machine-learned features extracted using FCN were combined with hand-crafted texton features to encode global information and local dependencies into feature representation. The score map with pixel-wise predictions was used as a feature map which was learned from multi-modal BRATS2017 training dataset using the FCN. The machine-learned features, along with hand-crafted texton features were then applied to random forests to classify each MRI image voxel.

The results of the FCN based method showed that the application of the RF classifier to multimodal MRI images using machine-learned features based on FCN and hand-crafted features based on textons provides promising segmentations. The mean Dice overlap measure for automatic brain tumor segmentation against ground truth is 0.86, 078 and 0.66 for the whole tumor, core and enhancing tumor, respectively.